\documentclass[11pt]{article}
\pdfoutput=1

\usepackage{microtype}
\usepackage{authblk}
\usepackage{url}
\usepackage[numbers]{natbib}
\usepackage[colorlinks,bookmarksopen,bookmarksnumbered,citecolor=red,urlcolor=red]{hyperref}

\begin{document}

\title{Users prefer Guetzli JPEG over same-sized libjpeg}
\author{J.~Alakuijala}
\author{R.~Obryk}
\author{Z.~Szabadka}
\author{\href{mailto:janwas@google.com}{J.~Wassenberg}}
\affil{Google Research}
\maketitle

\pdfinfo{
   /Author (J.~Alakuijala, R.~Obryk, Z.~Szabadka, J.~Wassenberg)
   /Title (Users prefer Guetzli JPEG over same-sized libjpeg)
   /CreationDate (D:20170309000000)
   /Subject ()
   /Keywords ()
   /Creator ()
   /Producer ()
}

\begin{abstract}

We report on pairwise comparisons by human raters of JPEG images from libjpeg
and our new Guetzli encoder. Although both files are size-matched, 75\% of
ratings are in favor of Guetzli. 
This implies the Butteraugli psychovisual image similarity metric which guides
Guetzli is reasonably close to human perception at high quality levels.
We provide access to the raw ratings and source images for further analysis
and study.
\end{abstract}

\section{Introduction}%
\label{sec:intro}%

Guetzli is a new JPEG encoder that uses the recently introduced Butteraugli
psychovisual similarity metric to make rate/distortion decisions. Thus,
Guetzli produces images that Butteraugli believes to be `better' than standard
libjpeg. We undertook an experiment to see whether humans agree with this
assessment. The settings are very simple: we have pairwise comparisons between
size-matched images, both in standard JPEG format, and see whether human
raters prefer one or the other.

\section{Materials and Methods}%
\label{sec:methods}%

\subsection{Source images}%

To ensure a meaningful evaluation of compressor quality and output size, we
create a 31-image dataset with known source and processing. The images are
publicly available \cite{imageDataset}.

We reduce JPEG artifacts by capturing images at the highest JPEG quality level
using a Canon EOS~600d camera and downsampling the resulting images by 4x4
using Lanczos resampling, as implemented in GIMP. Photographers often apply
unsharp masking to compensate for downsampling, so we also apply it in most of
the images before downsampling. The degree of unsharp masking is chosen
arbitrarily, but before any compression experiments. The images are chosen to
cover a wide range of contents, including nature, humans, smooth gradients,
high-frequency detail, with relatively thorough coverage of the sRGB gamut.

\subsection{Image degradation method}%

We are testing two compressors which offer a distortion vs size trade-off via
a single quality parameter. Guetzli is designed for high-quality, visually
lossless compression, so we choose its quality parameter to be 94, which
results in a rate of approximately 2.6~bits per pixel. For each
guetzli-compressed image, we generate a JPEG image for comparison by invoking
ImageMagick~6.7 via
\texttt{convert -sampling-factor 1x1} with decreasing quality parameter until
the libjpeg output is smaller than the guetzli JPEG. The final libjpeg JPEG
file is generated at the next higher quality level, which guarantees it is at
least as large as (and typically larger than) the Guetzli file. Note that the
scale of the resulting libjpeg quality parameters is slightly different; we
see a minimum of 83, maximum~93, average~89.4 and median~90. Both compressors
produce a standard-conformant JPEG bitstream.

Before display, we upsample both images by a factor of two using nearest
neighbor sampling (i.e. pixel replication) and crop them to 900x900~pixels
starting at the top-left corner so that the images fit on our screen.

\subsection{Viewing environment}%

To reduce the variability of the comparison results, e.g. due to differences
in monitor gamut, panel bit depth and processing/LUTs, we perform all tests on
a single monitor. A calibrated 27" NEC~PA272 includes a 10-bit panel while
still being reasonably commercially available. We choose viewing conditions
that match a typical office environment.

\subsection{Experiment design}%

We choose a pairwise comparison model. To avoid the need for a break, we
present the 31~images once in a single session, typically 20 to 30~minutes. We
provide a custom OpenGL viewer that alternates between the two compressed
variants on the right half of the screen, while displaying the uncompressed
original image on the left side. The presentation order of each image is
chosen randomly, and we swap between them at a fixed rate of 0.44~Hz. To reset
perception between the two images, we fade the screen to mid-gray over 250~ms
and hold it at gray for 600~ms. Our instructions are to click upon the less
preferable image, at the location of the most visible artifact. To reduce the
likelihood of random guesses, we leave open the option of skipping images
when there is no discernible difference.

Subjects are asked to sit at a distance of 3-4 picture heights from the screen
in order to reduce eye movements when comparing with the original image. We
provide a brief explanation and training, including the example of clicking on
a location.

\subsection{Subjects}%

23~raters participated in our experiment. Their age range spans 25-46~years
(median~31). 56.5\% (13) are women. All but three subjects report
corrected-to-normal vision; one is red-green color blind, one has slightly
higher visual acuity and one has slightly lower acuity. Subjects are recruited
via convenience sampling from Google employees working nearby, or known to us.
We attempt to equalize the gender ratio and include several experienced
photographers, but most subjects are not experienced in the field of image
compression and only informed that they are comparing two compressors.

\section{Results}%
\label{sec:results}%

Each subject generated 11 to 31~answers, with a mean of 26 and median of 29.
The raw ratings are listed below. Overall, 75\% of decisions were in favor of
Guetzli, i.e. the rater decided that the corresponding libjpeg image was
worse. There was only moderate variation among images; the interquartile range
is 85\%-68\%. In one extreme, only 22\% preferred the Guetzli encoding
of the `cloth' image, apparently due to loss of detail in the pink goggles.
Conversely, raters unanimously preferred the Guetzli encoding of `bees'.

The number of ratings for each image was relatively consistent (quartiles 19,
20, 21), which implies raters skipped different images. Assuming image
preferences are `uncertain' if the image was in the lower quartile of number
of ratings, discarding those images raises the median preference for Guetzli
to 80\%. Note that this analysis method was devised after the data was
collected, so it is possible the sampling method is biased towards a
particular outcome. Although discarding images that received the fewest
ratings seems reasonable, we must ignore this conclusion and only report the
75\% preference from the full dataset, listed below for completeness.

The ratings are listed in matrix form, one row per image, in decreasing order
of total rating decisions. Each column (corresponding to the rater's index)
indicates which is worse: L for libjpeg, G for Guetzli or blank if the image
was skipped.

\begin{verbatim}
 out-of-focus: LLLLLLGLLLLLLLLLGG LLLL
 white-yellow: LLLGLLGLL LLLLLG LLLLGG
  brake-light: LLLLLLLLG LGLGLLLG GL L
  pink-flower: LLLLLLLLL  LGLGG GLLLLL
     wollerau: LLLLGLGLLGLLLLLGLLLLLLL
  red-flowers: LLGGLGLLL LLL LLLL LLGG
    geranium2: LLLLGL LLLLLLGLGLLLLLLL
   green-rose: LGLLLLGGL G LLLG G LLLL
     bicycles: LLLLLLLLLGLLG GLL LLLLL
    blue-rose: LLLGLLGLLLG GLLL L LLLG
   pimpinelli: LLLGGL LL L LL GLLLLLLL
   minerology: LLLLLL LL L GLLG LLLLGG
     red-room: LLLLGLGLGL LL LL GLLLLL
      yellow2: LLLLLL LLGLLLGLL GLLLLL
         stp2: LLLLGLLLLLLLLLLLLLLLLLL
      vflower: LLLLLL LLLLLGLGG LLLLG 
         port: LLLLGLLLLGLLLLLGLL LLLL
      station: LGLGLLGGLLGGLLG    LLLL
       yellow: LGLGLLLGLGLLLLG   LLLLG
       bench2: LGLGGLLLLL  GLGL LGLLLG
     red-rose: LLGLLL LGL GGGGG L GL  
     geranium: LGLLGL LLGG GG L   GLLL
          stp: LGLLLLGLLGGLLGLL LLLLLL
      rainbow: GGGGLG LGLG G LL G GGLG
        bench: LLLGGLLLLL  GLGL L LLGG
          rgb: LLLLGG LLGLLLGL LLLLLLG
        cloth: GLGGGGGGLLLGG  G G GG G
       lichen: LGLGLLLLL GLG L  L  LGL
         bees: LLLLLLLLL LLL LL  LLLLL
        green: LLLLLL LL LLLLL    LLLG
         hand: GGL GL LL LGGLLLGL L LL
\end{verbatim}

\section{Discussion}%
\label{sec:discussion}%

Our headline result is that 75\% of mostly naive users (with no explicit
instructions about what kind of artifacts to look for) preferred the Guetzli
output over libjpeg, although both are in standard JPEG format and
size-matched (as far as possible using the JPEG quality parameter).

We created and published a new image dataset because the Kodak images are
low-resolution (512 $\times$ 768~pixels) and scanned from film
\cite{kodakStudy}. This leads to differences in color gamut, correlation,
saturation level and noise distribution versus today's digital cameras.
The IMAX/McMaster images are higher resolution, but apparently also scanned
from film \cite{mcmasterDataset}.

Our results are only valid for high-bitrate compression, which is useful for
long-term photo storage. Note that the recent ICIP 2016 Grand Challenge
\cite{icipChallenge}
included a subjective codec comparison using bitrates of up to 1.5~bits per
pixel, whereas our quality settings lead to about 2.6~bpp.
ITU~P910 \cite{itu910} recommends a darkened viewing environment to maximize
detection of artifacts, but we used normal office viewing conditions. This
more closely matches expected use, but may make artifacts harder to see.
However, we expect this to affect both
codecs equally. Absolute or degradation category ratings are unsuitable
measures because most of our subjects are naive, and the compression artifacts
are too minor to capture on a 5-point scale. Similarly, a Mean Opinion Score
may result in higher variability due to interpretation and/or language
differences \cite{mosLimitations}.
Hence, we only gathered pairwise comparisons.

Note that user preferences differ - some will be more sensitive to certain
kinds of artifacts, but the percentage of raters preferring Guetzli were
relatively consistent across images ($\pm 5\%$ from the median after
discarding the 25\% images with the fewest ratings).

In future studies of this kind, we suggest allowing users to toggle the two
compressor outputs at their discretion, because many felt that the current
rate was too fast. However, there should still be a certain upper bound on the
flicker rate to avoid highlighting compression artifacts via ``motion
detection''. Also, the statistical power would have been increased if we had
established rules prior to the data collection for discarding ratings.
Although ``outlier'' raters were not expected nor observed, most were untrained
and apparently struggled initially to detect compression artifacts. This leads
to some near-random ratings. One reasonable way to reduce these is to discard
images with the fewest votes. It would also be interesting to measure the
repeatability of observer ratings (i.e. noise level), and Type~I errors (false
alarms, i.e. detections of artifacts where none exist) by inserting some
identical image pairs.

In this study, we compared size-matched images. It might also be interesting
to determine how far Guetzli file sizes can be reduced until the rater
preference for Guetzli decreases to 50\%. This would indicate a ``compression
factor'' versus libjpeg, but reliably obtaining such a threshold would
be challenging given potentially noisy results.

\section{Conclusion}%
\label{sec:conclusion}%

In a pairwise comparison by 23 human raters of size-matched Guetzli vs libjpeg
encodings of 31~publicly available images, 75\% of all 614~comparisons were in
favor of Guetzli. We conclude that Guetzli generates higher quality images (at
considerable encoding cost).

This is due to the action of Butteraugli, which guides Guetzli.
We conclude that at very high qualities, the Butteraugli metric is
a reasonable proxy for human rater preference.

\bibliographystyle{unsrtnat}
\bibliography{references}{}

\end{document}